\documentclass[10pt,twocolumn,letterpaper]{article}

\usepackage{cvpr}              

\usepackage[dvipsnames]{xcolor}

\usepackage{amsmath,amssymb,amsfonts}
\usepackage{algorithm}
\usepackage{algpseudocode}
\usepackage{graphicx}
\usepackage{textcomp}
\usepackage{xcolor}
\usepackage{subcaption}
\usepackage{booktabs} 
\usepackage{tabularx}

\DeclareMathOperator*{\argmax}{argmax}

\usepackage{placeins}
\definecolor{cvprblue}{rgb}{0.21,0.49,0.74}
\usepackage[pagebackref,breaklinks,colorlinks,citecolor=cvprblue]{hyperref}

\def\paperID{40} 
\def\confName{CVPR}
\def\confYear{2024}

\title{Domain-independent detection of known anomalies}

\author{Jonas Bühler\\
\small Karlsruhe Institute of Technology (KIT)\\
\small \& preML GmbH\\
\small \href{https://preml.io/}{https://preml.io/}\\
{\tt\small jonas.buehler@preml.io}
\and
Jonas Fehrenbach\\
\small preML GmbH\\
\small \href{https://preml.io/}{https://preml.io/}\\
{\tt\small jonas.fehrenbach@preml.io}
\and
Lucas Steinmann\\
\small preML GmbH\\
\small \href{https://preml.io/}{https://preml.io/}\\
{\tt\small lucas.steinmann@preml.io}
\and
Christian Nauck\\
\small preML GmbH\\
\small \href{https://preml.io/}{https://preml.io/}\\
{\tt\small christian.nauck@preml.io}
\and
Marios Koulakis\\
\small Karlsruhe Institute of Technology (KIT)\\
\small \href{https://www.kit.edu/}{https://www.kit.edu/}\\
}

\usepackage{multirow}
\usepackage{svg}

\newcommand{\category}{\textit}
\newcommand{\class}{\textit}

\graphicspath{{images/}}

\begin{document}
    \maketitle
    
    \begin{abstract}
One persistent obstacle in industrial quality inspection is the detection of anomalies. In real-world use cases, two problems must be addressed: anomalous data is sparse and the same types of anomalies need to be detected on previously unseen objects.
Current anomaly detection approaches can be trained with sparse nominal data, whereas domain generalization approaches enable  detecting objects in previously unseen domains.
Utilizing those two observations, we introduce the hybrid task of domain generalization on sparse classes.
To introduce an accompanying dataset for this task, we present a modification of the well-established MVTec AD dataset by generating three new datasets.
In addition to applying existing methods for benchmark, we design two embedding-based approaches, Spatial Embedding MLP (SEMLP) and Labeled PatchCore. Overall, SEMLP achieves the best performance with an average image-level AUROC of 87.2~\% vs. 80.4~\% by MIRO. The new and openly available datasets allow for further research to improve industrial anomaly detection.
\end{abstract}

    \section{Introduction}
In industrial environments, machine-made components are susceptible to flaws and necessitate inspection before shipping. Typically, this task is performed manually by human operators who visually inspect each component. However, as with any repetitive task, human fatigue can lead to errors and the potential oversight of faulty components. Additionally, humans tend to judge anomalies inconsistently, with some being more stringent than others. Consequently, there is a growing interest in applying computer vision models to automate this process in a cost-effective manner while also reducing errors.

However, applying conventional classification approaches to this task introduces new challenges. First, anomalies may be rare, making it  difficult to collect a sufficient number of images for training. Second, not all possible types of anomalies may be known a priori, requiring models to be able to detect previously unseen anomaly types akin to humans capabilities. To address these challenges, anomaly detection methods were developed that can be trained effectively with limited training data. The advancement of robust anomaly detection models remains a subject of active research interest. For a comprehensive overview of industrial image anomaly detection, readers are directed to the reviews: \cite{taoDeepLearningUnsupervised2022,cuiSurveyUnsupervisedAnomaly2023,liuDeepIndustrialImage2024}.

Existing approaches have demonstrated the capacity to achieve near-perfect results on existing benchmark datasets \cite{rothTotalRecallIndustrial2022,yuFastFlowUnsupervisedAnomaly2021,hyunReConPatchContrastivePatch2024}. However, they are limited to detect anomalies, but cannot predict the type of the anomaly. In certain applications, there may be a necessity to differentiate between various types of anomalies in order to manage them accordingly. For instance, while some anomalies may be severe, others may be acceptable for a particular use case, such as color stains or flipped objects.

Another issue is that in certain industrial contexts, a company may produce a multitude of different objects \cite{liuDeepIndustrialImage2024}. Collecting sufficient data for all potential objects, even if it is merely normal images, would require a lot of time and resources. It would be beneficial if one could reuse a trained anomaly detection model for new objects. In addition to from the work presented in \Cref{s:related_work}, which have their own disadvantages, it is unclear whether existing anomaly detection models can be applied to previously unseen domains without retraining them. For future applications, models should work reliably on objects that are not known a priori.
Therefore, the first problem to solve is the lack of industrial anomaly detection datasets that cover multiple domains \cite{liuDeepIndustrialImage2024}.

Our main contributions are:
\begin{itemize}
    \item We propose the hybrid task of detecting known anomalies across different, previously unseen objects.
    \item We create three new datasets for this task, which are based on the popular MVTec Anomaly Detection dataset \cite{bergmannMVTecADComprehensive2019,bergmannMVTecAnomalyDetection2021}.
    \item We present two new methods which provide benchmarks on the the task on the modified MVTec datasets: Labeled PatchCore and Spatial Embedding MLP (SEMLP).
\end{itemize}

The paper is organized as follows: We commence with a review of related work (\Cref{s:related_work}). Subsequently, we detail the process of dataset generation and introduce the new methods, Labeled PatchCore and SEMLP (\Cref{s:methods}). Following this, \Cref{s:experiments-results} specifies the conducted experiments and presents the results. Finally, we draw conclusions (\Cref{s:conclusion}).

\section{Related Work}
\label{s:related_work}
Recently, many approaches were presented to solve the problem of anomaly detection. Furthermore, several domain generalization approaches tackle the problem of classifying previously unseen objects. We will give a quick overview over the recent approaches.
Furthermore, due to the lack of appropriate datasets for our tasks, we will also briefly describe existing datasets.

\subsection{Anomaly Detection}
    There are many different approaches to solve visual anomaly detection tasks on a single object using only normal images. We will present the two most common types of anomaly detection approaches in the first two paragraphs, representation-based and reconstruction-based ones, followed by some more exotic approaches.
    
    Representation-based approaches use a backbone model like ResNet \cite{heDeepResidualLearning2016} or one of its many extensions (e.g. \cite{zagoruykoWideResidualNetworks2017,xieAggregatedResidualTransformations2017a}) to create embeddings for single patches of the input image by extracting the outputs of intermediate layers. These patches can be used to learn a representation of normal patches. For example PaDiM \cite{defardPaDiMPatchDistribution2020} estimates their distribution and reports out-of-distribution test embeddings as anomalous. PatchCore \cite{rothTotalRecallIndustrial2022} creates a coreset which stores the embeddings of normal patches and uses the distance between test embeddings and the coreset as anomaly score.
    
    Reconstruction-based approaches train a model to reconstruct normal images, e.g. a (variational) autoencoder or generative adversarial network \cite{bergmannImprovingUnsupervisedDefect2024,schleglUnsupervisedAnomalyDetection2017}. Since anomalies are not present in the training data, it can be assumed, that these models fail to reconstruct anomalous areas. Hence, a pixel-based reconstruction error measure can be used as an anomaly score.
    
    Another approach, that does not fit into the categories above, is \cite{liCutPasteSelfSupervisedLearning2021}, which creates artificial anomalies to train a CNN as binary classifier, or few-shot anomaly detection approaches.
    
    Due to their simplicity, we will focus on representation-based approaches, especially PatchCore, which reports great results in the literature.

\subsection{Domain Generalization}
    Inspired by the fact that the domain of the dataset does not necessarily match the domains during the real-world usage, many papers have explored the problem of domain generalization \cite{zhouDomainGeneralizationSurvey2022}, where the domains of the training and test dataset differ.
    One common scenario is an autonomous car which must be able to detect and classify objects under all possible combinations of weather and environment conditions like illumination, fog, rain and snow \cite{chenDomainAdaptiveFaster2018,wuVectorDecomposedDisentanglementDomainInvariant2021}.
    
    \citet{chaDomainGeneralizationMutualInformation2022} proposes an approach called MIRO which applies a regularization loss while training a feature extractor. It uses a pre-trained model that has already seen many different types of images and applies an advanced fine-tuning. We will use this approach as one of the baselines.
    
    A first step towards domain generalization on the MVTec dataset is shown by \citet{chenDomainGeneralizedTexturedSurface2022}. It uses normal and anomalous images for training. But it has some limitations such as requiring normal images from the target domain during test time and only working on texture categories.
    
    \citet{huangRegistrationBasedFewShot2022} also explores anomaly detection in a domain generalization setting: While many approaches train their models for only a single category at a time, this approach trains a category-agnostic model with a few images from multiple categories. It can detect anomalies on new categories without requiring any retraining. However, it still needs some normal images from the new category during testing.

\subsection{Dataset}
    \label{s:related-work:datasets}
    The MVTec Anomaly Detection dataset \cite{bergmannMVTecADComprehensive2019,bergmannMVTecAnomalyDetection2021} introduced in 2019 is widely used in anomaly detection papers \cite{defardPaDiMPatchDistribution2020,rothTotalRecallIndustrial2022,yiPatchSVDDPatchlevel2020,liCutPasteSelfSupervisedLearning2021,yuFastFlowUnsupervisedAnomaly2021,liAnomalyDetectionSelforganizing2021} and quickly became a common benchmark dataset.
    It contains 15 different categories of which 10 are single objects and 5 textured surfaces. It covers dozens of different anomaly types for which it provides pixel-wise segmentation maps.
    Newer approaches \cite{rothTotalRecallIndustrial2022,yuFastFlowUnsupervisedAnomaly2021} already report nearly perfect results for classification and segmentation on it.
    
    A dataset used for domain generalization is PACS \cite{liDeeperBroaderArtier2017}, whose name is based on its 4 domains: photos, art paintings, cartoons, and sketches. It contains 7 different categories and a total of 9991 images.
    
    \citet{caoAnomalyDetectionDistribution2023} recently tried to adapt different datasets for anomaly detection with out-of-distribution test data. The PACS dataset is adapted for the anomaly detection part by defining one class as normal and all others as anomalous. The same is done for the MNIST dataset, where the images from the MNIST-M dataset are used as an additional domain. As for the MVTec dataset, new domains are created by augmenting the images.
    
    The new adapted datasets are a good first step towards domain generalization in anomaly detection, but as for the PACS and MNIST adaption, they use completely different objects as anomalies whereas anomalies are usually only small flaws on the normal object. The anomalous images of the MVTec adaption are out-of-distribution as advertised, but consist of only slight augmentations of the same object. None of these datasets allow to test the detection of small flaws on completely different objects.

    \begin{table*}[ht]
    \centering
    \caption{This table shows how anomaly types across different categories are merged into one dataset. Each category is treated as its own domain.
    For example the categories \category{cable}, \category{carpet}, \category{hazelnut}, \category{leather} \& \category{wood} have similar looking anomaly types called \class{hole} or \class{poke} which are used to create a new dataset called \textit{hole} (left column).}
    \begin{tabularx}{\linewidth}{p{2cm}X|p{2cm}X|p{2cm}X}
        \toprule
        \multicolumn{2}{c}{\class{hole} (\Cref{f:approach:domain-generalization:anomaly-types:hole})} & \multicolumn{2}{c}{\class{cut} (\Cref{f:approach:domain-generalization:anomaly-types:cut})} & \multicolumn{2}{c}{\class{color} (\Cref{f:approach:domain-generalization:anomaly-types:color})} \\
        \midrule
        category & anomaly type & category & anomaly type & category & anomaly type \\
        \midrule
        \category{cable} & \class{poke\_insulation} & \category{cable} & \class{cut\_inner\_insulation} & \category{carpet} & \class{color} \\
        \category{carpet} & \class{hole} & \category{cable} & \class{cut\_outer\_insulation} & \category{hazelnut} & \class{print} \\
        \category{hazelnut} & \class{hole} & \category{carpet} & \class{cut} & \category{leather} & \class{color} \\
        \category{leather} & \class{poke} & \category{hazelnut} & \class{cut} & \category{metal\_nut} & \class{color} \\
        \category{wood} & \class{hole} & \category{leather} & \class{cut} & \category{pill} & \class{color} \\
        & & \category{tile} & \class{crack} & \category{tile} & \class{gray\_stroke} \\
        & & & & \category{tile} & \class{oil} \\
        & & & & \category{wood} & \class{color} \\
        \bottomrule
    \end{tabularx}
    \label{t:approach:domain-generalization:categories}
\end{table*}

\section{Methods}
\label{s:methods}
First, we create our new datasets by modifying the MVTec AD dataset.
Then, we introduce two new approaches as baselines: i) we extend PatchCore by introducing multiple Corests, which we refer to as Labeled PatchCore, and ii) we apply a regular MLP to the embeddings used by PatchCore (Spatial Embedding MLP). The method is called spatial embedding, because the embeddings are extracted from vision ML models such as Transformers and CNNs where the embeddings contain spatially related information in contrast to MLPs.

\subsection{Generation of new datasets}
    For our task, we want to propose a dataset which contains multiple objects with similar anomalies. This would allow us to treat each object type (category) as another domain, so that we can train the detection of a specific anomaly on some of the categories and test it on previously unseen categories.
    As shown in \Cref{s:related-work:datasets}, none of the existing datasets satisfies our requirements. Therefore, three new datasets for domain generalized anomaly detection are created based on the MVTec AD dataset \cite{bergmannMVTecADComprehensive2019,bergmannMVTecAnomalyDetection2021}.
    
    For the first step, we look at all anomalies from all categories and determine which anomalies look similar.
    For example, the five categories (\category{cable}, \category{carpet}, \category{hazelnut}, \category{leather} \& \category{wood}) have anomaly types called either \class{hole} or \class{poke}, which we deem similar looking.
    
    In the next step, we create a dataset based on the categories containing these similar anomaly types. We use their good images and the anomalous images from said anomaly type.
    In our example, we would use the normal images from the previously mentioned categories (\category{cable}, \category{carpet}, \category{hazelnut}, \category{leather} \& \category{wood}) as well as the \class{hole} or \class{poke} anomaly images from these objects. We use all those images to create a new dataset called \textit{hole}.

    The same procedure is applied to create two more datasets, that we refer to as \textit{cut} and \textit{color}. Hence, we generate the three datasets: \textit{hole}, \textit{cut} and \textit{color}.
    Details regarding the exact object and corresponding anomaly types are listed for all three datasets in \Cref{t:approach:domain-generalization:categories}. To illustrate the objects and anomalies, example images are provided in the appendix (\Cref{f:dataset-examples}).

\subsection{Extension of PatchCore: Labeled PatchCore}
    Since we now have anomalous images available from the training domains, we aim to extend the idea of PatchCore \cite{rothTotalRecallIndustrial2022} to improve classification by correctly identifying the type of the anomaly.
    PatchCore collects all good embeddings in a coreset. We leverage the additional information from anomalous images by creating a second coreset for the embeddings from anomalous image parts (i.e. an anomaly coreset). The second coreset can be used to improve the results.
    More details can be found in \Cref{s:appendix:labeled-patchcore}.

\subsection{Simplified Approach: Spatial Embedding MLP}
\label{s:approach:classification:embedding-mlp}
    \begin{figure}
        \centering
        \includegraphics[width=.7\linewidth]{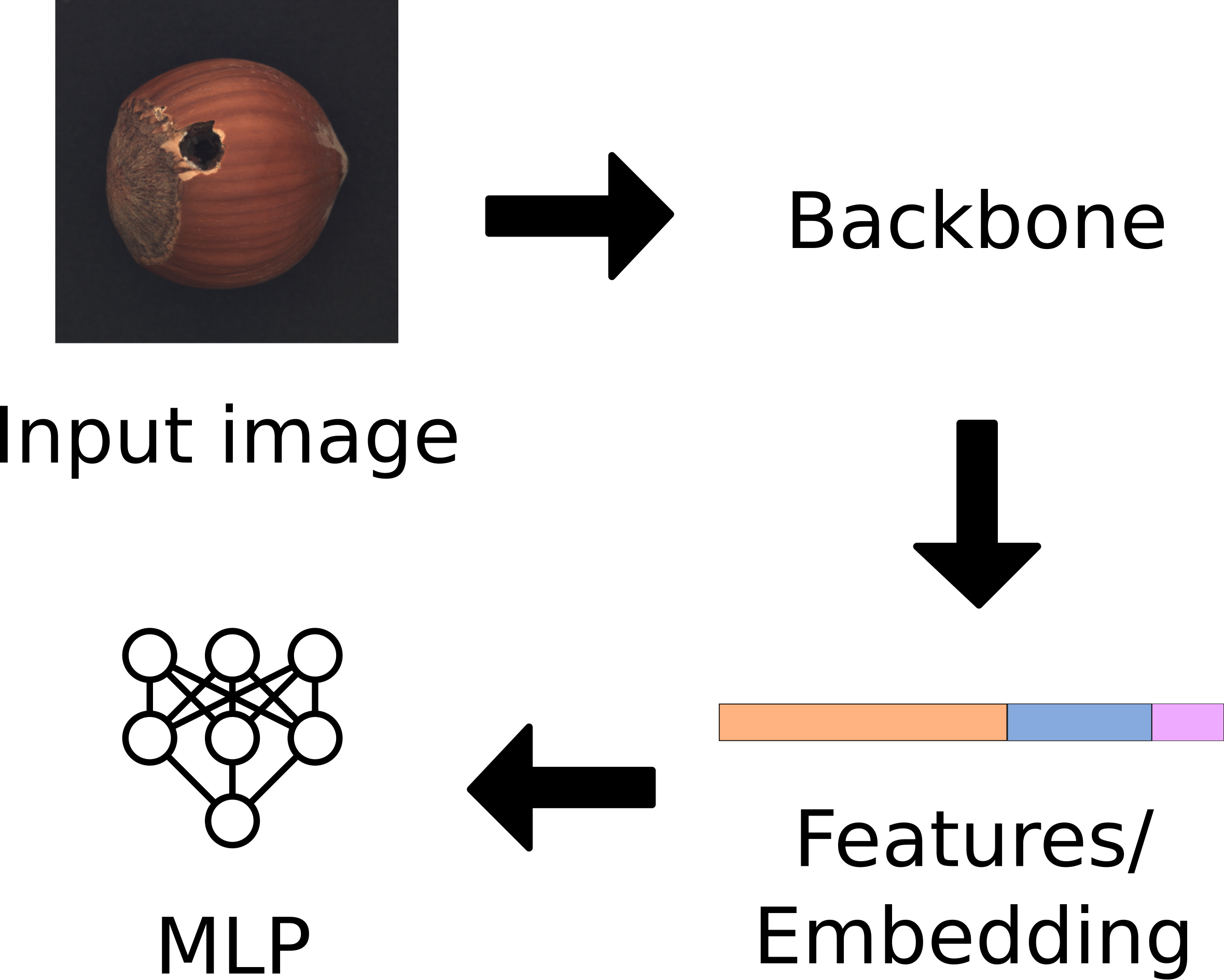}
        \caption{SEMLP passes the input image to a backbone. The intermediate features are extracted and concatenated to create embeddings. Each embedding is passed to an MLP which will classify single embeddings as normal or anomalous. The image is considered anomalous if at least one embedding is classified as anomalous.}
        \label{f:semlp-flow-chart}
    \end{figure}
    
    Instead of explicitly providing structure to the classification process by measuring the distances between the embeddings, we now analyze the capabilities of MLPs to directly classify the embeddings.
    For this approach, we extract embeddings, just like described above for the Labeled PatchCore. The embeddings are used as inputs to train small MLPs (about 49k trainable parameters), which are supposed to classify all single embeddings as good or anomalous. We call this approach Spatial Embedding MLP (SEMLP). \Cref{f:semlp-flow-chart} visualizes the classification flow.
    
    Training on patches gives us the advantage of having a lot more training data: While some anomaly classes contain less than a dozen images, we get a multitude of input data by training on patch embeddings. Even classes with only a few images have now about one hundred embeddings, some even over a thousand.

    \section{Experiments}

\label{s:experiments-results}

\begin{table*}[h]
    \centering
    \caption{Image-level AUROC in \% for all datasets with different approaches using Wide-Resnet50-2 and a Vision Transformer. (Note: the average is calculated across all target domains, as shown in \ref{t:approach:domain-generalization:resnet:auroc} and \ref{t:approach:domain-generalization:vit:auroc}.)}
    \footnotesize
    \begin{tabularx}{\linewidth}{l|XXXX|XXXX}
        \toprule
        & \multicolumn{3}{c}{Wide-ResNet50-2} & \multicolumn{3}{c}{ViT-B/8}\\
        & Cut & Color & Hole & Avg. & Cut & Color & Hole & Avg. \\
        \midrule
        SEMLP & \textbf{92.8}\tiny{ ± 2.5} & \textbf{82.9}\tiny{ ± 2.1} & 87.6\tiny{ ± 2.5} & \textbf{87.2} & \textbf{85.7}\tiny{ ± 3.9} & \textbf{85.7}\tiny{ ± 1.4} & \textbf{89.0}\tiny{ ± 3.2} & \textbf{86.7} \\
        PatchCore & 71.1\tiny{ ± 3.0} & 66.4\tiny{ ± 10.8} & 79.8\tiny{ ± 12.1} & 71.7 & 67.6\tiny{ ± 4.8} & 57.7\tiny{ ± 8.9} & 64.0\tiny{ ± 4.6} & 62.5 \\
        Labeled PatchCore & 47.3\tiny{ ± 2.3} & 48.4\tiny{ ± 3.9} & 46.2\tiny{ ± 2.2} & 47.4 & 47.7\tiny{ ± 2.1} & 56.8\tiny{ ± 2.0} & 48.7\tiny{ ± 1.5} & 51.7 \\
        MIRO & 87.0\tiny{ ± 2.8} & 71.5\tiny{ ± 4.2} & 86.9\tiny{ ± 2.3} & 80.4 & 75.9\tiny{ ± 8.2} & 75.4\tiny{ ± 5.1} & 79.1\tiny{ ± 4.3} & 76.6 \\
        SEMLP (MIRO) & 92.0\tiny{ ± 3.1} & 81.1\tiny{ ± 0.8} & \textbf{87.6}\tiny{ ± 1.8} & 86.2 & 77.4\tiny{ ± 5.8} & 80.0\tiny{ ± 5.1} & 79.3\tiny{ ± 7.4} & 79.0 \\
        PatchCore (MIRO) &  67.6\tiny{ ± 0.5} & 69.5\tiny{ ± 3.0} & 73.9\tiny{ ± 4.4} & 70.2 & 68.1\tiny{ ± 1.9} & 80.9\tiny{ ± 7.5} & 71.7\tiny{ ± 4.0} & 74.4 \\
        \bottomrule
    \end{tabularx}
    \label{t:results:auroc}
\end{table*}

\subsection{Setup}
\label{s:experiments}
To evaluate our proposed models as well as the baselines at detecting anomalies across different categories, we always choose one category from the dataset as target domain and use the other ones as source domains for training. The anomalous images of the source domains are split evenly into training and validation set, whereas we use all anomalous images from the target domain for testing.

As backbone models for SEMLP, we use \mbox{Wide-ResNet50-2} and Vision Transformers (ViT). Details can be found in \Cref{s:appendix:backbones}. We compare the methods to PatchCore (cf.~\Cref{s:appendix:patchcore-details} and MIRO \cite{chaDomainGeneralizationMutualInformation2022}. For MIRO, we continue training on the pre-trained \mbox{Wide-ResNet50-2} and ViT respectively. We report those results as plain "MIRO". Furthermore, we use the adjusted models as backbones to extract embeddings for SEMLP and PatchCore (which we report as "SEMLP (MIRO)" and "PatchCore (MIRO)").

    \subsection{Results}
\label{s:results}
To evaluate the performance of the tested models and allow for comparison runs, we provide the resulting image-level AUROC for Wide-Resnet50-2 and Vision Transformer as backbones in \Cref{t:results:auroc}. More detailed results for the image-level AUROC and F1-Score can be found in the appendix in \Cref{t:results:f1score,t:approach:domain-generalization:resnet:auroc,t:approach:domain-generalization:vit:auroc}.
As we can see, SEMLP achieves an image-level AUROC of 87.2~\% with the WideResNet backbone and is already nearly perfect on many categories. Using the trained models from MIRO as backbones for SEMLP does not work better then regular backbones pretrained on ImageNet.

Even though SEMLP achieves good average performances, SEMLP fails pretty badly on some categories (like the \textit{pill} in the \textit{color} dataset or the \textit{cable} in the \textit{hole} dataset), even though other methods perform well. Future work can identify the reasons, potential limitations and provide improvement strategies to increase the generalization performance. Plain MIRO seems to be the second best approach, followed by plain PatchCore, which is better than our Labeled PatchCore approach. Providing the additional structure in our way is not helpful to improve the performance of PatchCore. It would be interesting to study if other modifications can further increase the performance of PatchCore.

Furthermore, the anomaly scores of some target domains are significantly higher than for the source domains. But since the anomalous images usually have an even higher score, there still exists a good threshold in most cases.

Even though our approaches require no data from the target domain to train the models, one limitation of our method is the need for normal and anomalous images from the target domain to determine a good threshold.

    \section{Conclusion}
\label{s:conclusion}
We presented a new task on domain generalization of sparse anomalous classes.
To conduct this task, we introduced three new datasets based on the MVTec AD dataset to compare different anomaly detection models on domain generalization tasks.

We presented first baselines on these datasets, covering some existing models and introducing two new approaches.
One of this is SEMLP which classifies single embeddings and requires only a few labeled training images. The new model can distinguish normal and anomalous data in most cases and achieves an image AUROC of 87.2~\%, which is better than MIRO.
Notably SEMLP outperforms PatchCore and also Labeled PatchCore. Apparently providing additional structure by introducing the labeled coresets hinders prediction and it is beneficial to purely rely on the predictive performance of MLPs.

Even though our approach does not require large amounts of data, we still need some labeled images of the new domain to determine a good threshold. Although a threshold can be easily set by the user, future work may focus on eliminating this need by finding a good threshold. Nevertheless, SEMLP shows potential for real-world applications. Especially the opportunity of detecting known anomalies can be advantageous for industrial applications.

\FloatBarrier

    \bibliographystyle{ieeenat_fullname}
    \bibliography{main}

\begin{thebibliography}{31}
\providecommand{\natexlab}[1]{#1}
\providecommand{\url}[1]{\texttt{#1}}
\expandafter\ifx\csname urlstyle\endcsname\relax
  \providecommand{\doi}[1]{doi: #1}\else
  \providecommand{\doi}{doi: \begingroup \urlstyle{rm}\Url}\fi

\bibitem[Akcay et~al.(2018)Akcay, Atapour-Abarghouei, and Breckon]{akcayGANomalySemiSupervisedAnomaly2018}
Samet Akcay, Amir Atapour-Abarghouei, and Toby~P. Breckon.
\newblock {GANomaly}: {Semi}-{Supervised} {Anomaly} {Detection} via {Adversarial} {Training}, 2018.
\newblock arXiv:1805.06725 [cs].

\bibitem[Akcay et~al.(2022)Akcay, Ameln, Vaidya, Lakshmanan, Ahuja, and Genc]{akcayAnomalibDeepLearning2022}
Samet Akcay, Dick Ameln, Ashwin Vaidya, Barath Lakshmanan, Nilesh Ahuja, and Utku Genc.
\newblock Anomalib: {A} {Deep} {Learning} {Library} for {Anomaly} {Detection}, 2022.
\newblock arXiv:2202.08341 [cs].

\bibitem[Bergmann et~al.(2019)Bergmann, Fauser, Sattlegger, and Steger]{bergmannMVTecADComprehensive2019}
Paul Bergmann, Michael Fauser, David Sattlegger, and Carsten Steger.
\newblock {MVTec} {AD} — {A} {Comprehensive} {Real}-{World} {Dataset} for {Unsupervised} {Anomaly} {Detection}.
\newblock In \emph{2019 {IEEE}/{CVF} {Conference} on {Computer} {Vision} and {Pattern} {Recognition} ({CVPR})}, pages 9584--9592, 2019.
\newblock ISSN: 2575-7075.

\bibitem[Bergmann et~al.(2021)Bergmann, Batzner, Fauser, Sattlegger, and Steger]{bergmannMVTecAnomalyDetection2021}
Paul Bergmann, Kilian Batzner, Michael Fauser, David Sattlegger, and Carsten Steger.
\newblock The {MVTec} {Anomaly} {Detection} {Dataset}: {A} {Comprehensive} {Real}-{World} {Dataset} for {Unsupervised} {Anomaly} {Detection}.
\newblock \emph{International Journal of Computer Vision}, 129\penalty0 (4):\penalty0 1038--1059, 2021.

\bibitem[Bergmann et~al.(2024)Bergmann, Löwe, Fauser, Sattlegger, and Steger]{bergmannImprovingUnsupervisedDefect2024}
Paul Bergmann, Sindy Löwe, Michael Fauser, David Sattlegger, and Carsten Steger.
\newblock Improving {Unsupervised} {Defect} {Segmentation} by {Applying} {Structural} {Similarity} to {Autoencoders}.
\newblock pages 372--380, 2024.

\bibitem[Cao et~al.(2023)Cao, Zhu, and Pang]{caoAnomalyDetectionDistribution2023}
Tri Cao, Jiawen Zhu, and Guansong Pang.
\newblock Anomaly {Detection} under {Distribution} {Shift}, 2023.
\newblock arXiv:2303.13845 [cs].

\bibitem[Cha et~al.(2022)Cha, Lee, Park, and Chun]{chaDomainGeneralizationMutualInformation2022}
Junbum Cha, Kyungjae Lee, Sungrae Park, and Sanghyuk Chun.
\newblock Domain {Generalization} by {Mutual}-{Information} {Regularization} with {Pre}-trained {Models}, 2022.
\newblock arXiv:2203.10789 [cs].

\bibitem[Chen et~al.(2022)Chen, Liu, Lin, Chen, and Wang]{chenDomainGeneralizedTexturedSurface2022}
Shang-Fu Chen, Yu-Min Liu, Chia-Ching Lin, Trista Pei-Chun Chen, and Yu-Chiang~Frank Wang.
\newblock Domain-{Generalized} {Textured} {Surface} {Anomaly} {Detection}, 2022.
\newblock arXiv:2203.12304 [cs].

\bibitem[Chen et~al.(2018)Chen, Li, Sakaridis, Dai, and Van~Gool]{chenDomainAdaptiveFaster2018}
Yuhua Chen, Wen Li, Christos Sakaridis, Dengxin Dai, and Luc Van~Gool.
\newblock Domain {Adaptive} {Faster} {R}-{CNN} for {Object} {Detection} in the {Wild}, 2018.
\newblock arXiv:1803.03243 [cs].

\bibitem[Cui et~al.(2023)Cui, Liu, and Lian]{cuiSurveyUnsupervisedAnomaly2023}
Yajie Cui, Zhaoxiang Liu, and Shiguo Lian.
\newblock A {Survey} on {Unsupervised} {Anomaly} {Detection} {Algorithms} for {Industrial} {Images}.
\newblock \emph{IEEE Access}, 11:\penalty0 55297--55315, 2023.
\newblock arXiv:2204.11161 [cs].

\bibitem[Defard et~al.(2020)Defard, Setkov, Loesch, and Audigier]{defardPaDiMPatchDistribution2020}
Thomas Defard, Aleksandr Setkov, Angelique Loesch, and Romaric Audigier.
\newblock {PaDiM}: a {Patch} {Distribution} {Modeling} {Framework} for {Anomaly} {Detection} and {Localization}, 2020.
\newblock arXiv:2011.08785 [cs].

\bibitem[Dosovitskiy et~al.(2021)Dosovitskiy, Beyer, Kolesnikov, Weissenborn, Zhai, Unterthiner, Dehghani, Minderer, Heigold, Gelly, Uszkoreit, and Houlsby]{dosovitskiyImageWorth16x162021}
Alexey Dosovitskiy, Lucas Beyer, Alexander Kolesnikov, Dirk Weissenborn, Xiaohua Zhai, Thomas Unterthiner, Mostafa Dehghani, Matthias Minderer, Georg Heigold, Sylvain Gelly, Jakob Uszkoreit, and Neil Houlsby.
\newblock An {Image} is {Worth} 16x16 {Words}: {Transformers} for {Image} {Recognition} at {Scale}, 2021.
\newblock arXiv:2010.11929 [cs].

\bibitem[{Falcon, William}()]{Pytorchlightning}
{Falcon, William}.
\newblock pytorch-lightning https://github.com/{Lightning}-{AI}/pytorch-lightning/.

\bibitem[He et~al.(2016)He, Zhang, Ren, and Sun]{heDeepResidualLearning2016}
Kaiming He, Xiangyu Zhang, Shaoqing Ren, and Jian Sun.
\newblock Deep {Residual} {Learning} for {Image} {Recognition}.
\newblock In \emph{2016 {IEEE} {Conference} on {Computer} {Vision} and {Pattern} {Recognition} ({CVPR})}, pages 770--778, 2016.
\newblock ISSN: 1063-6919.

\bibitem[Huang et~al.(2022)Huang, Guan, Jiang, Zhang, Spratling, and Wang]{huangRegistrationBasedFewShot2022}
Chaoqin Huang, Haoyan Guan, Aofan Jiang, Ya Zhang, Michael Spratling, and Yan-Feng Wang.
\newblock Registration based {Few}-{Shot} {Anomaly} {Detection}, 2022.
\newblock arXiv:2207.07361 [cs].

\bibitem[Hyun et~al.(2024)Hyun, Kim, Jeon, Kim, Bae, and Kang]{hyunReConPatchContrastivePatch2024}
Jeeho Hyun, Sangyun Kim, Giyoung Jeon, Seung~Hwan Kim, Kyunghoon Bae, and Byung~Jun Kang.
\newblock {ReConPatch} : {Contrastive} {Patch} {Representation} {Learning} for {Industrial} {Anomaly} {Detection}, 2024.
\newblock arXiv:2305.16713 [cs].

\bibitem[Li et~al.(2021{\natexlab{a}})Li, Sohn, Yoon, and Pfister]{liCutPasteSelfSupervisedLearning2021}
Chun-Liang Li, Kihyuk Sohn, Jinsung Yoon, and Tomas Pfister.
\newblock {CutPaste}: {Self}-{Supervised} {Learning} for {Anomaly} {Detection} and {Localization}, 2021{\natexlab{a}}.
\newblock arXiv:2104.04015 [cs].

\bibitem[Li et~al.(2017)Li, Yang, Song, and Hospedales]{liDeeperBroaderArtier2017}
Da Li, Yongxin Yang, Yi-Zhe Song, and Timothy~M. Hospedales.
\newblock Deeper, {Broader} and {Artier} {Domain} {Generalization}, 2017.
\newblock arXiv:1710.03077 [cs].

\bibitem[Li et~al.(2021{\natexlab{b}})Li, Jiang, Ma, Wei, Hong, and Gong]{liAnomalyDetectionSelforganizing2021}
Ning Li, Kaitao Jiang, Zhiheng Ma, Xing Wei, Xiaopeng Hong, and Yihong Gong.
\newblock Anomaly {Detection} via {Self}-organizing {Map}, 2021{\natexlab{b}}.
\newblock arXiv:2107.09903 [cs].

\bibitem[Liu et~al.(2024)Liu, Xie, Wang, Li, Wang, Zheng, and Jin]{liuDeepIndustrialImage2024}
Jiaqi Liu, Guoyang Xie, Jinbao Wang, Shangnian Li, Chengjie Wang, Feng Zheng, and Yaochu Jin.
\newblock Deep {Industrial} {Image} {Anomaly} {Detection}: {A} {Survey}.
\newblock \emph{Machine Intelligence Research}, 21\penalty0 (1):\penalty0 104--135, 2024.
\newblock arXiv:2301.11514 [cs].

\bibitem[Paszke et~al.(2019)Paszke, Gross, Massa, Lerer, Bradbury, Chanan, Killeen, Lin, Gimelshein, Antiga, Desmaison, Kopf, Yang, DeVito, Raison, Tejani, Chilamkurthy, Steiner, Fang, Bai, and Chintala]{paszkePyTorchImperativeStyle2019}
Adam Paszke, Sam Gross, Francisco Massa, Adam Lerer, James Bradbury, Gregory Chanan, Trevor Killeen, Zeming Lin, Natalia Gimelshein, Luca Antiga, Alban Desmaison, Andreas Kopf, Edward Yang, Zachary DeVito, Martin Raison, Alykhan Tejani, Sasank Chilamkurthy, Benoit Steiner, Lu Fang, Junjie Bai, and Soumith Chintala.
\newblock {PyTorch}: {An} {Imperative} {Style}, {High}-{Performance} {Deep} {Learning} {Library}.
\newblock In \emph{Advances in {Neural} {Information} {Processing} {Systems} 32}, pages 8024--8035. Curran Associates, Inc., 2019.

\bibitem[Roth et~al.(2022)Roth, Pemula, Zepeda, Schölkopf, Brox, and Gehler]{rothTotalRecallIndustrial2022}
Karsten Roth, Latha Pemula, Joaquin Zepeda, Bernhard Schölkopf, Thomas Brox, and Peter Gehler.
\newblock Towards {Total} {Recall} in {Industrial} {Anomaly} {Detection}, 2022.
\newblock arXiv:2106.08265 [cs].

\bibitem[Schlegl et~al.(2017)Schlegl, Seeböck, Waldstein, Schmidt-Erfurth, and Langs]{schleglUnsupervisedAnomalyDetection2017}
Thomas Schlegl, Philipp Seeböck, Sebastian~M. Waldstein, Ursula Schmidt-Erfurth, and Georg Langs.
\newblock Unsupervised {Anomaly} {Detection} with {Generative} {Adversarial} {Networks} to {Guide} {Marker} {Discovery}, 2017.
\newblock arXiv:1703.05921 [cs].

\bibitem[Tao et~al.(2022)Tao, Gong, Zhang, Yan, and Adak]{taoDeepLearningUnsupervised2022}
Xian Tao, Xinyi Gong, Xin Zhang, Shaohua Yan, and Chandranath Adak.
\newblock Deep {Learning} for {Unsupervised} {Anomaly} {Localization} in {Industrial} {Images}: {A} {Survey}.
\newblock \emph{IEEE Transactions on Instrumentation and Measurement}, 71:\penalty0 1--21, 2022.
\newblock Conference Name: IEEE Transactions on Instrumentation and Measurement.

\bibitem[Wightman(2019)]{rw2019timm}
Ross Wightman.
\newblock {PyTorch} image models, https://github.com/rwightman/pytorch-image-models, 2019.

\bibitem[Wu et~al.(2021)Wu, Liu, Han, Zhu, and Yang]{wuVectorDecomposedDisentanglementDomainInvariant2021}
Aming Wu, Rui Liu, Yahong Han, Linchao Zhu, and Yi Yang.
\newblock Vector-{Decomposed} {Disentanglement} for {Domain}-{Invariant} {Object} {Detection}, 2021.
\newblock arXiv:2108.06685 [cs].

\bibitem[Xie et~al.(2017)Xie, Girshick, Dollár, Tu, and He]{xieAggregatedResidualTransformations2017a}
Saining Xie, Ross Girshick, Piotr Dollár, Zhuowen Tu, and Kaiming He.
\newblock Aggregated {Residual} {Transformations} for {Deep} {Neural} {Networks}, 2017.
\newblock arXiv:1611.05431 [cs].

\bibitem[Yi and Yoon(2020)]{yiPatchSVDDPatchlevel2020}
Jihun Yi and Sungroh Yoon.
\newblock Patch {SVDD}: {Patch}-level {SVDD} for {Anomaly} {Detection} and {Segmentation}, 2020.
\newblock arXiv:2006.16067 [cs].

\bibitem[Yu et~al.(2021)Yu, Zheng, Wang, Li, Wu, Zhao, and Wu]{yuFastFlowUnsupervisedAnomaly2021}
Jiawei Yu, Ye Zheng, Xiang Wang, Wei Li, Yushuang Wu, Rui Zhao, and Liwei Wu.
\newblock {FastFlow}: {Unsupervised} {Anomaly} {Detection} and {Localization} via {2D} {Normalizing} {Flows}, 2021.
\newblock arXiv:2111.07677 [cs].

\bibitem[Zagoruyko and Komodakis(2017)]{zagoruykoWideResidualNetworks2017}
Sergey Zagoruyko and Nikos Komodakis.
\newblock Wide {Residual} {Networks}, 2017.
\newblock arXiv:1605.07146 [cs].

\bibitem[Zhou et~al.(2022)Zhou, Liu, Qiao, Xiang, and Loy]{zhouDomainGeneralizationSurvey2022}
Kaiyang Zhou, Ziwei Liu, Yu Qiao, Tao Xiang, and Chen~Change Loy.
\newblock Domain {Generalization}: {A} {Survey}.
\newblock \emph{IEEE Transactions on Pattern Analysis and Machine Intelligence}, pages 1--20, 2022.
\newblock arXiv:2103.02503 [cs].

\end{thebibliography}
    
    \clearpage
    \setcounter{section}{0}
    \renewcommand{\thesection}{\Alph{section}}
    \section{Appendix}

\subsection{Labeled PatchCore}
\label{s:appendix:labeled-patchcore}
    This section provides a more in-depth description of our Labeled PatchCore approach.
    
    Since anomalous images also contain normal areas, we will only consider patches of the anomaly itself for our anomaly coreset.
    Depending on the original resolution and chosen backbone, a single patch contains roughly 1024 (32x32) pixels of the original image.
    Therefore, we consider patches anomalous, if at least a tenth of the pixels are anomalous. This way, we obtain a sufficient share of anomalous patches for our anomaly coreset, without considering patches anomalous that just have one or two anomalous pixels. The good patches from anomalous images are discarded because there are already enough \class{good} patches from the normal images.
    
    Then, we add the embeddings of these patches (also called locally aware patch features in \cite{rothTotalRecallIndustrial2022}) to the memory bank of the anomaly coreset.
    Because of the limited number of anomalous images and patches, we always keep at least 1,000 embeddings for the anomaly coreset when subsampling or use all embeddings if necessary, regardless of the given coreset subsampling ratio.

    \begin{figure}
        \centering
        \includegraphics[width=.8\linewidth]{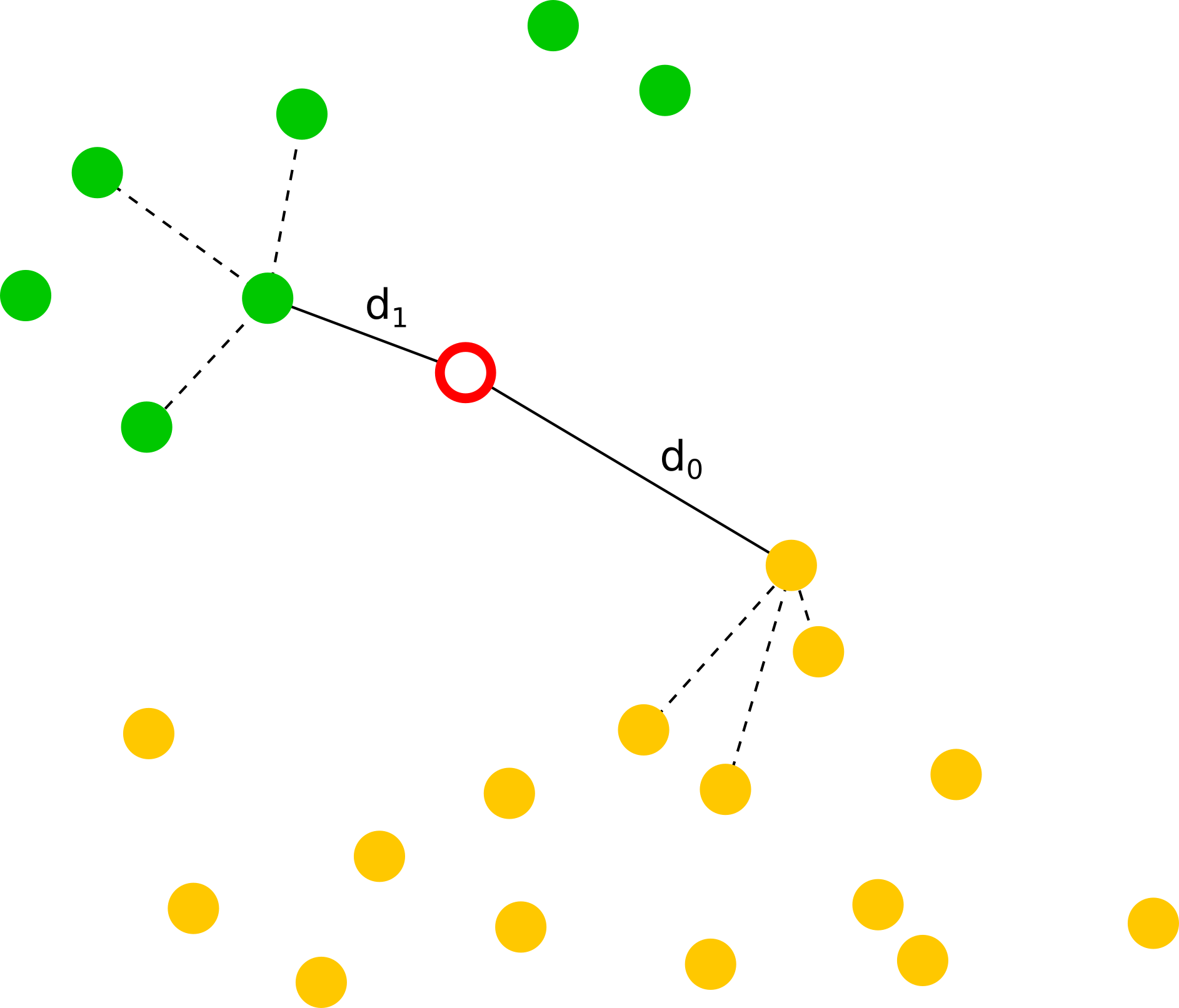}
        \caption{Classification using multiple coresets in a two-dimensional scenario: yellow is the coreset containing embeddings of good images and green the coreset with anomalous embeddings. The red circle will be classified based on its distance to the next embedding of each coreset ($d_0$ and $d_1$), after weighting the distances to the next neighbors in their respective coreset like in \citet{rothTotalRecallIndustrial2022} (dotted lines).}
        \label{fig:labeled-coreset}
    \end{figure}

    To classify an image, we extract the embeddings for each patch. Then, we classify each embedding based on whether they are closer to the normal or anomalous coreset. We hope that the second coreset provides more information and a better structure to differentiate the anomalies from normal patches.
    
    To do so, we calculate the weighted anomaly score for each embedding like the original implementation (see section 3.3 in \citet{rothTotalRecallIndustrial2022}) but in relation to each of the two coresets (as shown in \Cref{fig:labeled-coreset} for a two-dimensional example).
    In other words, each embedding is classified by calculating the distance $d_i$ to the next embedding $m_i^*$ for each coreset $M_i$. The distances are weighted with the nearest neighbors of $m_i^*$ in its coreset to get score $s_i$ (see Equation 7 in \cite{rothTotalRecallIndustrial2022}).
    Afterwards, we classify each patch as belonging to the coreset $M_i$ with the smallest score $s_i$, i.e. the coreset to which the embedding is closest after weighting.

    The process is repeated for all patches of the images. If all patches are classified as \class{good}, the image is deemed as normal. Otherwise it is considered anomalous.

\subsection{Details for experimental setup}
As in \citet{rothTotalRecallIndustrial2022}, we also resize the images to 256x256 and center crop them to 226x226 pixels and we use a neighbourhood size of 3 (i. e. apply an 3x3-average pooling across the patch features before creating the embeddings).

\subsubsection{Backbone models}
\label{s:appendix:backbones}
Similar to \citet{rothTotalRecallIndustrial2022}, we use \mbox{Wide-ResNet50-2} as backbone for all our tasks and use the features of the second and third block to create our embeddings.
Since Vision Transformers (ViT) have recently gained popularity in computer vision, we also test a ViT as backbone. In this case, we use the base variant as presented in \citet{dosovitskiyImageWorth16x162021} with a patch size of 8. Here, we extract the features after the layers 5 and 9. We use pre-trained weights provided by \citet{rw2019timm} for all backbones.

\subsubsection{Details on the application of the specific approaches}
\label{s:appendix:patchcore-details}
Unless otherwise mentioned, the SEMLP has two fully-connected layers with a hidden size of 32 and a leaky ReLU layer (with $\alpha = 0.01$) in-between and predicts the scalar anomaly score, which leads to relatively small MLPs with about 49k trainable parameters.

For comparison, we use the default PatchCore implementation, which only uses normal images to create a single coreset.
Because we now have a multitude of training data, we reduce the coreset subsampling rate for default PatchCore and Labeled PatchCore to 0.1 divided by the number of different categories in the current dataset (i.e. $\frac{0.1}{5}$ for \textit{hole} and \textit{cut}, and $\frac{0.1}{7}$ for \textit{color}) due to memory constraints. For this reason, we also implement an online subsampling algorithm for most categories, which returned nearly identical results on the default MVTec AD dataset (see \Cref{s:appendix:online-subsampling} for details).

\subsection{Online subsampling}
\label{s:appendix:online-subsampling}
    The default coreset subsampling algorithm of PatchCore collects all embeddings before creating the coreset. This will result in huge memory usage if using a larger amount of images or higher resolutions. It also requires that all training data is available before calculating the coreset and running any tests.
    
    We therefore define a simple algorithm to learn online, which allows us to calculate the coreset with minimal memory usage and to continue training at any time. This is done by making two small changes to the original subsampling algorithm in \citet{rothTotalRecallIndustrial2022}:
    First, we do not collect the embeddings beforehand but sample them as we create the coreset.
    Second, while PatchCore subsamples a fix percentage $r$ of all embeddings for the coreset, we add $r\%$ of the current image's embeddings to the coreset.
    Apart from that, the algorithm remains the same.
    
    The new algorithm is shown in \Cref{alg:online-coreset} whereas all inputs are defined exactly as in \citet{rothTotalRecallIndustrial2022}.

    We compared the default and the online subsampling algorithm on the default MVTec AD dataset and achieved nearly identical results (both having an average image-level AUROC of 98.7~\%).
    
    \begin{algorithm}
        \textbf{Input:}\par
            \hskip\algorithmicindent $\phi$: pre-trained feature extractor\par
            \hskip\algorithmicindent $j \in \mathbb{N}^n_0$: one or multiple indices of the layers from which the features are extracted\par
            \hskip\algorithmicindent $\mathcal{X}^{(train)}_N = \{\mathbf{x}_i | \mathbf{x}_i\in\mathbb{R}^{m\times n\times c}\}_{i\leq N}$: normal training data\par
            \hskip\algorithmicindent $P$: locally aware patch-feature collector\par
            \hskip\algorithmicindent $s \in \mathbb{N}$: stride (on patch-level after extracting the features, usually 1)\par
            \hskip\algorithmicindent $p \in \mathbb{N}$: neighbourhood size (kernel size of the average pooling, default: 3)\par
            \hskip\algorithmicindent $r \in (0, 1]$: coreset ratio (ratio of embeddings that are added to the coreset)\par
            \hskip\algorithmicindent $\psi$: random linear projection \\
        \textbf{Output:}\par
            \hskip\algorithmicindent $M_C$ memory bank
        \begin{algorithmic}[1]
            \State $M_C \gets \{\}$
            \For{$x_i \in \mathcal{X}^{(train)}_N$}
                \State $M \gets P_{s, p}(\phi_j(x_i))$  \label{alg:online-coreset:patch-sampling}
                \For{$i \in [0, ..., \lceil{|M| * r}\rceil - 1]$}
                    \State $m_i \gets \argmax\limits_{m \in M-M_C} \min\limits_{n \in M_C} \|\psi(m) - \psi(n)\|_2$
                    \State $M_C \gets M_C \cup \{m_i\}$
                \EndFor
            \EndFor
        \end{algorithmic}
        \caption{online coreset subsampling}
        \label{alg:online-coreset}
    \end{algorithm}
    
    \begin{algorithm}
        \textbf{Input:}\par
            \hskip\algorithmicindent $\phi$: pre-trained feature extractor\par
            \hskip\algorithmicindent $j \in \mathbb{N}^n_0$: one or multiple indices of the layers from which the features are extracted\par
            \hskip\algorithmicindent $\mathcal{X}^{(train)}_{N_b} = \{\mathbf{x}_i | \mathbf{x}_i\in\mathbb{R}^{m\times n\times c}\}_{i\leq N}$: normal training data\par
            \hskip\algorithmicindent $P$: locally aware patch-feature collector\par
            \hskip\algorithmicindent $s \in \mathbb{N}$: stride (on patch-level after extracting the features, usually 1)\par
            \hskip\algorithmicindent $p \in \mathbb{N}$: neighbourhood size (kernel size of the average pooling, default: 3)\par
            \hskip\algorithmicindent $r \in (0, 1]$: coreset ratio (ratio of embeddings that are added to the coreset)\par
            \hskip\algorithmicindent $\psi$: random linear projection\par
            \hskip\algorithmicindent $b \in \mathbb{N}$: batchsize \\
        \textbf{Output:}\par
            \hskip\algorithmicindent $M_C$ memory bank
        \begin{algorithmic}[1]
            \State $M_C \gets \{\}$
            \For{$(B := (x_k, x_{k+1}, ..., x_{k+b})) \in \mathcal{X}^{(train)}_{N_b}$}
                \State $M \gets \{\}$
                \For{$x_i \in B$}
                    \State $M \gets M \cup P_{s, p}(\phi_j(x_i))$
                \EndFor
                \For{$i \in [0, ..., \lceil{|M| * r}\rceil - 1]$}
                    \State $m_i \gets \argmax\limits_{m \in M-M_C} \min\limits_{n \in M_C} \|\psi(m) - \psi(n)\|_2$
                    \State $M_C \gets M_C \cup \{m_i\}$
                \EndFor
            \EndFor
        \end{algorithmic}
        \caption{batch-wise online coreset subsampling}
        \label{alg:online-coreset:batch}
    \end{algorithm}
    
    It is also possible to learn batch-wise by iterating over batches instead of images and simply accumulating all patches from all images of the batch instead of just one image in line \cref{alg:online-coreset:patch-sampling}.
    
    One could also think of more elaborate online learning algorithms, for example changing the number of added embeddings depending on how different the embeddings of the new image/batch are. But our tests show that our simple algorithm already achieves good results compared to the original one.

\subsection{Software details}
    To perform the experiments in this paper, we use anomalib \cite{akcayAnomalibDeepLearning2022}, a library that implements many state of the art anomaly detection models like PaDiM \cite{defardPaDiMPatchDistribution2020} and PatchCore \cite{rothTotalRecallIndustrial2022}, but also Fastflow \cite{yuFastFlowUnsupervisedAnomaly2021}, GANomaly \cite{akcayGANomalySemiSupervisedAnomaly2018} and others. It is based on PyTorch \cite{paszkePyTorchImperativeStyle2019} and PyTorchLightning \cite{Pytorchlightning}. It contains dataloaders for the MVTec AD dataset and allows an easy validation and visualization of the results. It also provides ready-to-use scripts for training and testing as well as configuration files.
    
    Anomalib uses PyTorch Image Models (timm) \cite{rw2019timm} which provides many ImageNet pre-trained models among other things. For example, it provides many different ResNet and Vision Transformer models.
    
    We will extend and adjust quite a few parts of the library to implement our experiments.

\subsection{Data and source code availability}
    The extended anomalib code can be found at \url{https://doi.org/10.5281/zenodo.11924708}.
    The dataset is available at \url{https://doi.org/10.5281/zenodo.11920346}.

\subsection{Dataset example images}
\begin{figure}
       \centering
       \begin{subfigure}{\linewidth}
           \centering
           \includegraphics[width=\linewidth]{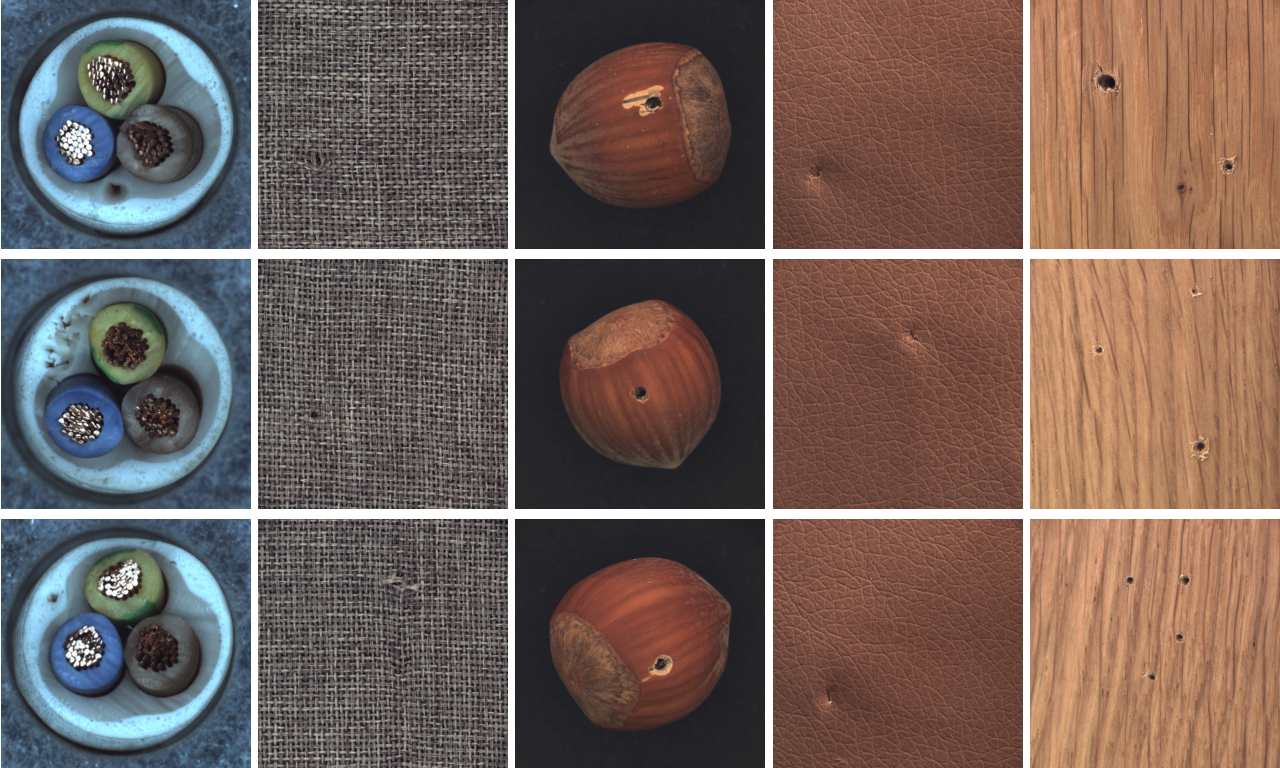}
           \caption{\textit{hole}}
           \label{f:approach:domain-generalization:anomaly-types:hole}
       \end{subfigure}
       \begin{subfigure}{\linewidth}
           \centering
           \includegraphics[width=\linewidth]{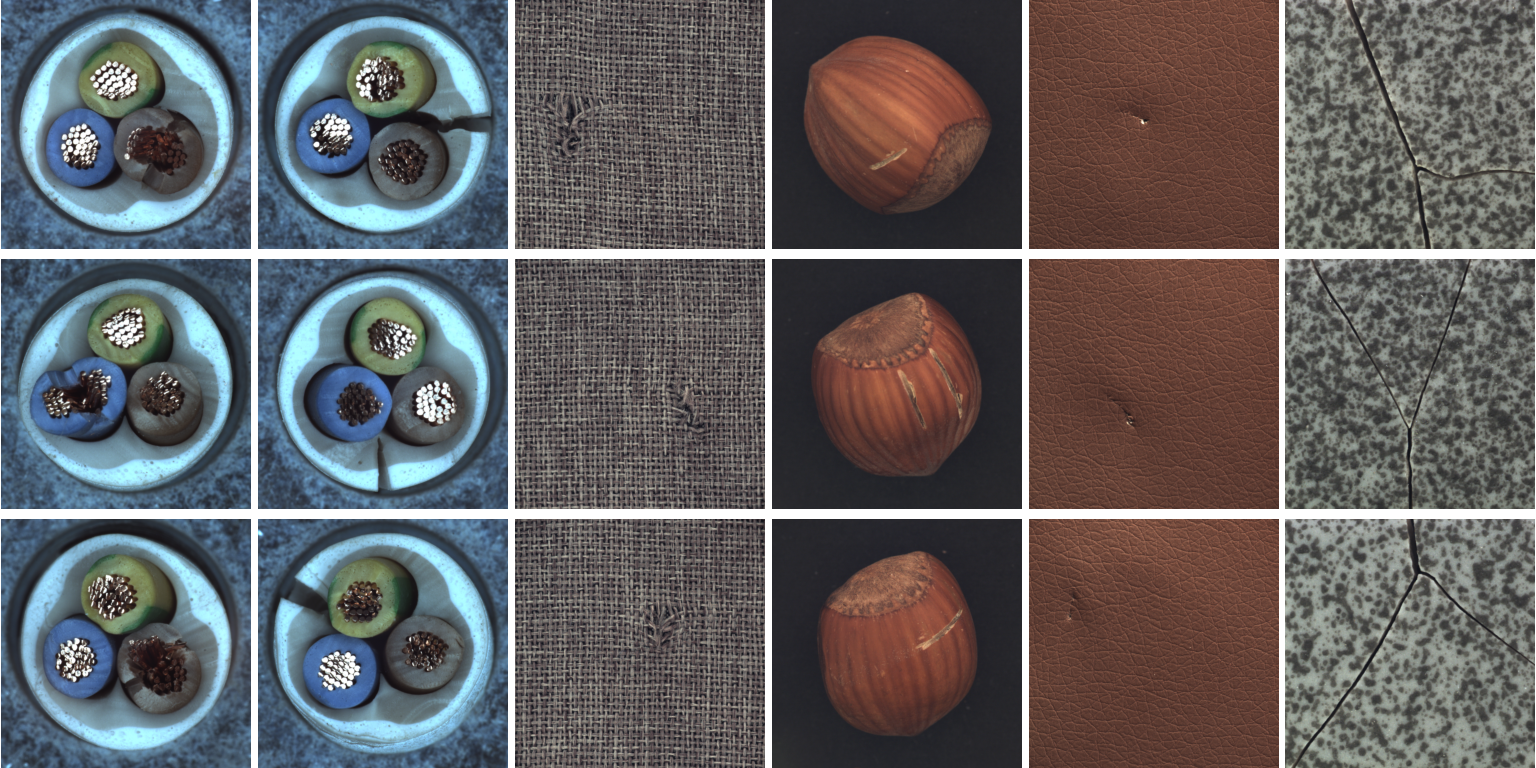}
           \caption{\textit{cut}}
           \label{f:approach:domain-generalization:anomaly-types:cut}
       \end{subfigure}
       \begin{subfigure}{\linewidth}
           \centering
           \includegraphics[width=\linewidth]{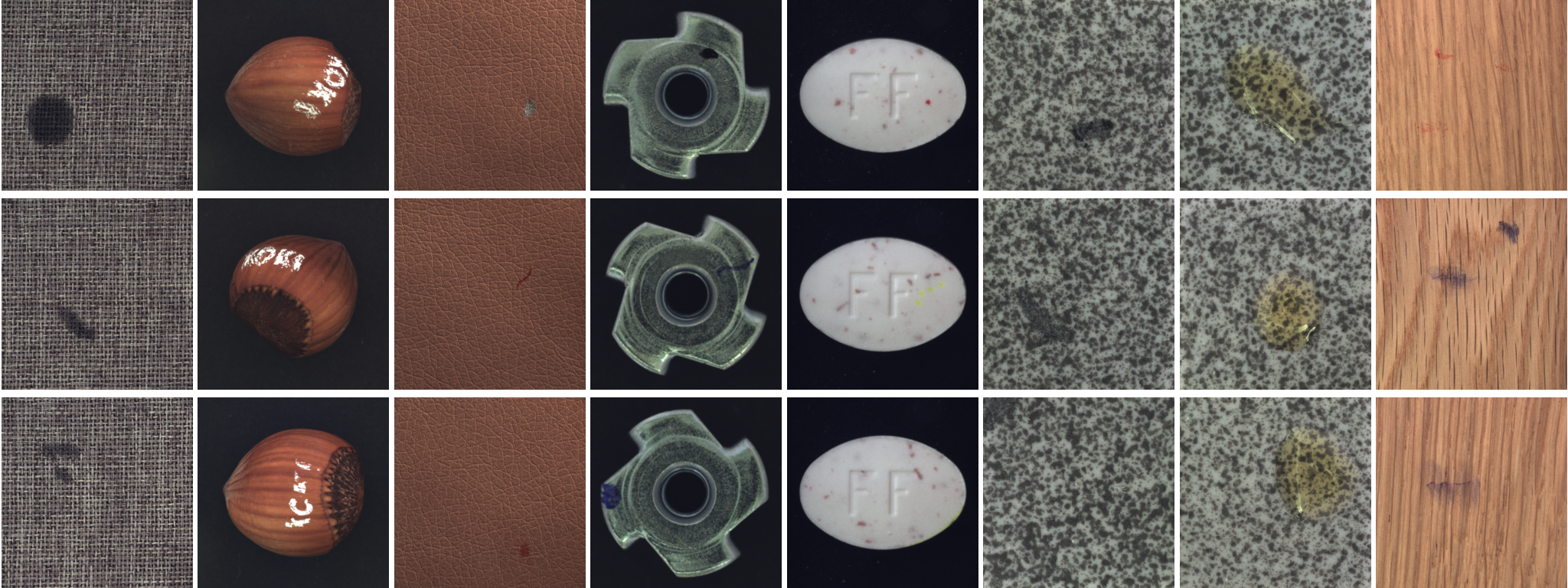}
           \caption{\textit{color}}
           \label{f:approach:domain-generalization:anomaly-types:color}
       \end{subfigure}
       \caption{Example images of the custom defined anomaly types (source of individual images: \cite{bergmannMVTecADComprehensive2019,bergmannMVTecAnomalyDetection2021})}
       \label{f:dataset-examples}
    \end{figure}

    \Cref{f:dataset-examples} shows some example images for each dataset.
    
\subsection{Detailed results}
    The following \Cref{t:results:f1score,t:approach:domain-generalization:resnet:auroc,t:approach:domain-generalization:vit:auroc} will show more detailed results from our experiments.

    \begin{table*}[h]
        \centering
        \caption{Image-level F1-Scores in \% for all datasets with different approaches using Wide-Resnet50-2 and a Vision Transformer}
        \footnotesize
        \begin{tabularx}{\linewidth}{l|XXXX|XXXX}
            \toprule
            & \multicolumn{4}{c}{Wide-ResNet50-2} & \multicolumn{4}{c}{ViT-B/8}\\
            & Cut & Color & Hole & Avg. & Cut & Color & Hole & Avg. \\
            \midrule
            SEMLP & 86.3\tiny{ ± 2.6} & \textbf{83.7} \tiny{ ± 1.0} & 83.2\tiny{ ± 1.7} & \textbf{84.3} & \textbf{82.3}\tiny{ ± 3.8} & \textbf{87.1}\tiny{ ± 1.3} & \textbf{84.9}\tiny{ ± 1.9} & \textbf{85.0} \\
            PatchCore & 66.4\tiny{ ± 3.4} & 70.9\tiny{ ± 7.4} & 72.4\tiny{ ± 16.2} & 70.0 & 66.6\tiny{ ± 5.6} & 64.9\tiny{ ± 4.7} & 58.0\tiny{ ± 3.0} & 63.4 \\
            Labeled PatchCore & 52.9\tiny{ ± 1.0} & 59.6\tiny{ ± 0.6} & 48.9\tiny{ ± 0.8} & 54.5 & 52.2\tiny{ ± 0.5} & 67.4\tiny{ ± 1.0} & 54.6\tiny{ ± 1.4} & 59.2 \\
            MIRO & 85.2\tiny{ ± 2.0} & 75.2\tiny{ ± 2.2} & 82.8\tiny{ ± 2.2} & 80.4 & 71.9\tiny{ ± 6.8} & 76.1\tiny{ ± 4.8} & 71.5\tiny{ ± 4.1} & 73.5 \\
            SEMLP (MIRO) & \textbf{87.7}\tiny{ ± 4.4} & 80.2 \tiny{ ± 0.8} & \textbf{84.3}\tiny{ ± 0.7} & 83.6 & 72.0\tiny{ ± 6.3} & 78.7\tiny{ ± 3.7} & 73.0\tiny{ ± 4.8} & 75.0 \\
            PatchCore (MIRO) & 65.6\tiny{ ± 1.1} & 71.0\tiny{ ± 1.8} & 65.7\tiny{ ± 3.7} & 67.9 & 65.1\tiny{ ± 2.9} & 80.0\tiny{ ± 6.5} & 66.2\tiny{ ± 3.9} & 71.6 \\
            \bottomrule
        \end{tabularx}
        \label{t:results:f1score}
    \end{table*}

    \begin{table*}
        \centering
        \footnotesize
        \caption{Image-level AUROC in \% for different target domains classified by different approaches using Wide-Resnet50-2}
        \begin{tabularx}{\linewidth}{llXXXXXX}
            \toprule
            Dataset & Target Domain & SEMLP & PatchCore & Labeled \mbox{PatchCore} & MIRO & SEMLP (MIRO) & PatchCore (MIRO) \\
            \midrule
            \multirow{5}{*}{Cut} & Cable & 76.0\tiny{ ± 11.0} & 63.5\tiny{ ± 1.5} & 59.6\tiny{ ± 13.6} & 47.9\tiny{ ± 16.8} & 73.6\tiny{ ± 10.6} & 67.1\tiny{ ± 2.6} \\
            & Carpet & 99.7\tiny{ ± 0.4} & 74.2\tiny{ ± 7.0} & 51.5\tiny{ ± 6.7} & 95.2\tiny{ ± 4.2} & 98.8\tiny{ ± 2.1} & 74.5\tiny{ ± 8.8} \\
            & Hazelnut & 91.0\tiny{ ± 6.2} & 68.6\tiny{ ± 3.1} & 46.0\tiny{ ± 2.4} & 99.1\tiny{ ± 1.6} & 98.0\tiny{ ± 2.1} & 60.9\tiny{ ± 3.2} \\
            & Leather & 98.6\tiny{ ± 3.0} & 49.0\tiny{ ± 5.7} & 55.0\tiny{ ± 7.1} & 93.1\tiny{ ± 6.9} & 90.4\tiny{ ± 9.4} & 36.2\tiny{ ± 11.9} \\
            & Tile & 98.5\tiny{ ± 1.9} & 100\tiny{ ± 0.0} & 24.6\tiny{ ± 11.8} & 99.4\tiny{ ± 0.3} & 99.0\tiny{ ± 1.2} & 99.4\tiny{ ± 0.5} \\
            \hline
            \multirow{7}{*}{Color} & Carpet & 100\tiny{ ± 0.0} & 67.0\tiny{ ± 8.5} & 55.0\tiny{ ± 12.7} & 85.5\tiny{ ± 20.2} &98.7\tiny{ ± 2.9} & 74.7\tiny{ ± 8.4} \\
            & Hazelnut & 53.4\tiny{ ± 9.3} & 86.9\tiny{ ± 1.5} & 34.5\tiny{ ± 5.2} & 52.1\tiny{ ± 11.8} & 60.0\tiny{ ± 7.4} & 80.4\tiny{ ± 16.2} \\
            & Leather & 99.2\tiny{ ± 1.5} & 35.4\tiny{ ± 1.8} & 59.0\tiny{ ± 8.0} & 82.6\tiny{ ± 17.0} & 90.6\tiny{ ± 13.1} & 55.0\tiny{ ± 17.1} \\
            & Metal Nut & 70.2\tiny{ ± 3.6} & 48.6\tiny{ ± 2.6} & 56.2\tiny{ ± 2.6} & 53.6\tiny{ ± 7.8} & 71.1\tiny{ ± 10.1} & 53.9\tiny{ ± 5.9} \\
            & Pill & 60.7\tiny{ ± 10.5} & 56.9\tiny{ ± 5.1} & 57.8\tiny{ ± 13.0} & 48.7\tiny{ ± 7.8} & 63.1\tiny{ ± 11.6} & 59.7\tiny{ ± 11.8} \\
            & Tile & 96.8\tiny{ ± 0.9} & 70.0\tiny{ ± 4.6} & 42.7\tiny{ ± 5.0} & 78.1\tiny{ ± 5.3} & 84.6\tiny{ ± 7.3} & 64.8\tiny{ ± 11.1} \\
            & Wood & 100\tiny{ ± 0.0} & 100\tiny{ ± 0.0} & 33.4\tiny{ ± 7.0} & 100\tiny{ ± 0.0} & 100\tiny{ ± 0.0} & 97.9\tiny{ ± 2.4} \\
            \hline
            \multirow{5}{*}{Hole} & Cable & 45.9\tiny{ ± 9.6} & 50.8\tiny{ ± 1.7} & 51.1\tiny{ ± 3.6} & 52.1\tiny{ ± 11.9} & 44.4\tiny{ ± 10.1} & 51.6\tiny{ ± 2.2} \\
            & Carpet & 97.6\tiny{ ± 1.8} & 87.9\tiny{ ± 3.3} & 49.0\tiny{ ± 3.3} & 98.0\tiny{ ± 2.5} & 98.6\tiny{ ± 1.5} & 80.4\tiny{ ± 6.1} \\
            & Hazelnut & 98.6\tiny{ ± 2.0} & 85.5\tiny{ ± 2.0} & 51.1\tiny{ ± 7.5} & 96.0\tiny{ ± 4.2} & 100.0\tiny{ ± 0.0} & 59.3\tiny{ ± 16.8} \\
            & Leather & 100\tiny{ ± 0.0} & 74.8\tiny{ ± 14.5} & 47.4\tiny{ ± 4.8} & 94.7\tiny{ ± 7.7} & 99.6\tiny{ ± 0.9} & 79.7\tiny{ ± 12.3} \\
            & Wood & 95.9\tiny{ ± 2.1} & 100\tiny{ ± 0.0} & 32.5\tiny{ ± 5.1} & 93.8\tiny{ ± 3.8} & 95.5\tiny{ ± 3.7} & 98.3\tiny{ ± 2.1} \\
            \hline
            \multicolumn{2}{c}{Average} & 87.2 & 71.7 & 47.4 & 80.6 & 86.2 & 70.2 \\
            \bottomrule
        \end{tabularx}
        \label{t:approach:domain-generalization:resnet:auroc}
    \end{table*}
    
    \begin{table*}
        \centering
        \footnotesize
        \caption{Image-level AUROC in \% for different target domains classified by different approaches using ViT-B/8}
        \begin{tabularx}{\linewidth}{llXXXXXX}
            \toprule
            Dataset & Target Domain & SEMLP & PatchCore & Labeled \mbox{PatchCore} & MIRO & SEMLP (MIRO) & PatchCore (MIRO) \\
            \midrule
            \multirow{5}{*}{Cut} & Cable & 52.3\tiny{ ± 8.9} & 69.7\tiny{ ± 2.8} & 57.8\tiny{ ± 2.1} & 56.0\tiny{ ± 4.3} & 56.5\tiny{ ± 4.0} & 65.0\tiny{ ± 3.3} \\
            & Carpet & 100\tiny{ ± 0.0} & 52.3\tiny{ ± 3.1} & 39.7\tiny{ ± 1.2} & 74.8\tiny{ ± 27.2} & 78.0\tiny{ ± 18.9} & 63.9\tiny{ ± 14.0} \\
            & Hazelnut & 77.8\tiny{ ± 18.0} & 35.9\tiny{ ± 1.8} & 32.0\tiny{ ± 2.8} & 73.3\tiny{ ± 15.0} & 73.0\tiny{ ± 13.2} & 58.1\tiny{ ± 22.5} \\
            & Leather & 100\tiny{ ± 0.0} & 90.2\tiny{ ± 1.5} & 43.7\tiny{ ± 5.2} & 85.5\tiny{ ± 8.8} & 93.2\tiny{ ± 4.6} & 74.4\tiny{ ± 17.8} \\
            & Tile & 98.6\tiny{ ± 1.4} & 90.2\tiny{ ± 1.8} & 65.0\tiny{ ± 1.9} & 89.9\tiny{ ± 11.9} & 86.4\tiny{ ± 15.2} & 79.1\tiny{ ± 25.4} \\
            \midrule
            \multirow{7}{*}{Color} & Carpet & 100\tiny{ ± 0.0} & 67.5\tiny{ ± 0.8} & 48.8\tiny{ ± 4.2} & 63.8\tiny{ ± 13.5} & 70.4\tiny{ ± 6.4} & 80.0\tiny{ ± 4.2} \\
            & Hazelnut & 84.8\tiny{ ± 18.3} & 36.0\tiny{ ± 1.6} & 98.4\tiny{ ± 0.7} & 69.1\tiny{ ± 16.9} & 67.4\tiny{ ± 20.3} & 76.7\tiny{ ± 13.1} \\
            & Leather & 100\tiny{ ± 0.0} & 28.9\tiny{ ± 4.7} & 49.4\tiny{ ± 8.9} & 92.4\tiny{ ± 17.0} & 90.5\tiny{ ± 21.3} & 85.0\tiny{ ± 31.1} \\
            & Metal Nut & 70.2\tiny{ ± 3.6} & 54.8\tiny{ ± 1.0} & 33.5\tiny{ ± 8.1} & 65.8\tiny{ ± 17.5} & 67.9\tiny{ ± 22.1} & 59.0\tiny{ ± 14.2} \\
            & Pill & 59.8\tiny{ ± 13.5} & 45.0\tiny{ ± 2.2} & 51.1\tiny{ ± 3.5} & 71.2\tiny{ ± 15.9} & 73.2\tiny{ ± 11.9} & 75.8\tiny{ ± 9.7} \\
            & Tile & 98.3\tiny{ ± 0.7} & 88.7\tiny{ ± 3.5} & 54.3\tiny{ ± 2.6} & 68.0\tiny{ ± 21.3} & 91.8\tiny{ ± 11.7} & 94.3\tiny{ ± 2.4} \\
            & Wood & 100\tiny{ ± 0.0} & 82.6\tiny{ ± 8.0} & 62.2\tiny{ ± 8.7} & 97.4\tiny{ ± 5.2} & 98.7\tiny{ ± 2.9} & 95.1\tiny{ ± 10.9} \\
            \midrule
            \multirow{5}{*}{Hole} & Cable & 47.1\tiny{ ± 16.3} & 44.3\tiny{ ± 2.1} & 80.4\tiny{ ± 2.2} & 54.2\tiny{ ± 10.0} & 55.2\tiny{ ± 15.6} & 51.8\tiny{ ± 11.0} \\
            & Carpet & 100\tiny{ ± 0.0} & 45.1\tiny{ ± 3.3} & 40.3\tiny{ ± 1.1} & 71.4\tiny{ ± 10.4} & 64.4\tiny{ ± 14.4} & 56.1\tiny{ ± 15.5} \\
            & Hazelnut & 100\tiny{ ± 0.0} & 71.0\tiny{ ± 5.2} & 7.1\tiny{ ± 1.5} & 92.3\tiny{ ± 15.5} & 98.2\tiny{ ± 2.4} & 82.8\tiny{ ± 15.5} \\
            & Leather & 100\tiny{ ± 0.0} & 83.0\tiny{ ± 3.2} & 54.2\tiny{ ± 1.2} & 81.0\tiny{ ± 9.5} & 82.7\tiny{ ± 24.0} & 84.1\tiny{ ± 16.8} \\
            & Wood & 97.8\tiny{ ± 1.0} & 76.5\tiny{ ± 6.1} & 61.6\tiny{ ± 5.1} & 96.7\tiny{ ± 5.3} & 96.1\tiny{ ± 3.6} & 83.5\tiny{ ± 22.7} \\
            \midrule
            \multicolumn{2}{c}{Average} & 86.7 & 62.5 & 51.7 & 76.6 & 79.0 & 74.4 \\
            \bottomrule
        \end{tabularx}
        \label{t:approach:domain-generalization:vit:auroc}
    \end{table*}

\end{document}